\def\year{2018}\relax
\documentclass[letterpaper]{article} 
\usepackage{aaai18}  
\usepackage{times}  
\usepackage{helvet}  
\usepackage{courier}  
\usepackage{url}  
\usepackage{graphicx}  
\usepackage{multirow}
\usepackage{subcaption}
\usepackage{tabularx}
\usepackage{amsmath}
\usepackage{amssymb}
\DeclareMathOperator{\K}{K}
\DeclareMathOperator{\N}{N}
\newcommand{\Ek}[2]{\ensuremath{{\mathbf{e}_{{#1}}^{{#2}}}}}
\newcommand{\Vk}[3]{\ensuremath{{\mathbf{#1}_{{#2}}^{{#3}}}}}

\usepackage{xspace}
\makeatletter
\DeclareRobustCommand\onedot{\futurelet\@let@token\@onedot}
\def\@onedot{\ifx\@let@token.\else.\null\fi\xspace}

\def\ie{\emph{i.e}\onedot} 
 
 \def\vs{\emph{vs}\onedot}

\makeatother
\usepackage{pgfplots}
\pgfplotsset{compat=1.3}
\pgfplotsset{
  /pgfplots/ybar legend/.style={
    /pgfplots/legend image code/.code={%
       \draw[##1,/tikz/.cd,yshift=-0.25em]
        (0cm,0cm) rectangle (3pt,0.8em);},
  },
  bar group size/.style 2 args={
      /pgf/bar shift={%
              -0.5*(#2*\pgfplotbarwidth + (#2-1)*\pgfkeysvalueof{/pgfplots/bar group skip})  +
              (.5+#1)*\pgfplotbarwidth + #1*\pgfkeysvalueof{/pgfplots/bar group skip}},%
  },
  bar group skip/.initial=2pt,
  plot 0/.style={blue,fill=blue!30!white,mark=none},%
  plot 1/.style={red,fill=red!30!white,mark=none},%
  plot 2/.style={teal!60!black,fill=teal!30!white,mark=none},%
  plot 3/.style={orange!60!black,fill=orange!30!white,mark=none},%
  plot 4/.style={black!60!black,fill=black!30!white,mark=none},%
}
\nocopyright
\begin{document}
%
\title{Compatibility Family Learning for Item Recommendation and Generation}
\author{
Yong-Siang Shih$^1$, Kai-Yueh Chang$^1$, Hsuan-Tien Lin$^{1,2}$, Min Sun$^3$\\
$^1$Appier Inc., Taipei, Taiwan,\\
$^2$National Taiwan University, Taipei, Taiwan,\\
$^3$National Tsing Hua University, Hsinchu, Taiwan.\\
\{yongsiang.shih,kychang\}@appier.com, htlin@csie.ntu.edu.tw, sunmin@ee.nthu.edu.tw
}
\maketitle
\begin{abstract}

Compatibility between items, such as clothes and shoes, is a major factor among customer's purchasing decisions.
However, learning ``compatibility" is challenging due to (1) broader notions of compatibility than those of similarity, (2) the asymmetric nature of compatibility, and (3) only a small set of compatible and incompatible items are observed.
We propose an end-to-end trainable system to embed each item into a latent vector and project a query item into $\K$ compatible
prototypes in the same space. These prototypes reflect the broad notions of compatibility. We refer to both the embedding and prototypes as ``Compatibility Family''.
In our learned space, we introduce a novel Projected Compatibility Distance (PCD) function which is
differentiable and ensures diversity by aiming for at least one prototype to be close
to a compatible item, whereas none of the prototypes are close to an incompatible item.
We evaluate our system on a toy dataset,
two Amazon product datasets, and Polyvore outfit dataset.
Our method consistently achieves state-of-the-art performance.
Finally, we show that we can visualize the candidate compatible prototypes using a Metric-regularized Conditional
Generative Adversarial Network (MrCGAN), where the input is a projected prototype and the output is
a generated image of a compatible item.
We ask human evaluators to judge the relative compatibility between our generated
images and images generated by CGANs conditioned directly on query items.
Our generated images are significantly preferred, with roughly twice the number of votes as others.
\end{abstract}

\section{Introduction}\label{sec.intro}
Identifying compatible items is an important aspect in building recommendation systems.
For instance, recommending matching shoes to a specific dress is important for fashion; recommending
a wine to go with different dishes is important for restaurants.
In addition, it is valuable to visualize what style is missing from the existing dataset so as to foresee potential matching items that could have been up-sold to the users. We believe that the generated compatible items could inspire fashion designers to create novel products and help our business clients to fulfill the needs of customers.

\begin{figure}
  \centering
   \begin{tabular}{c|c}
    \includegraphics[width=0.275\textwidth]{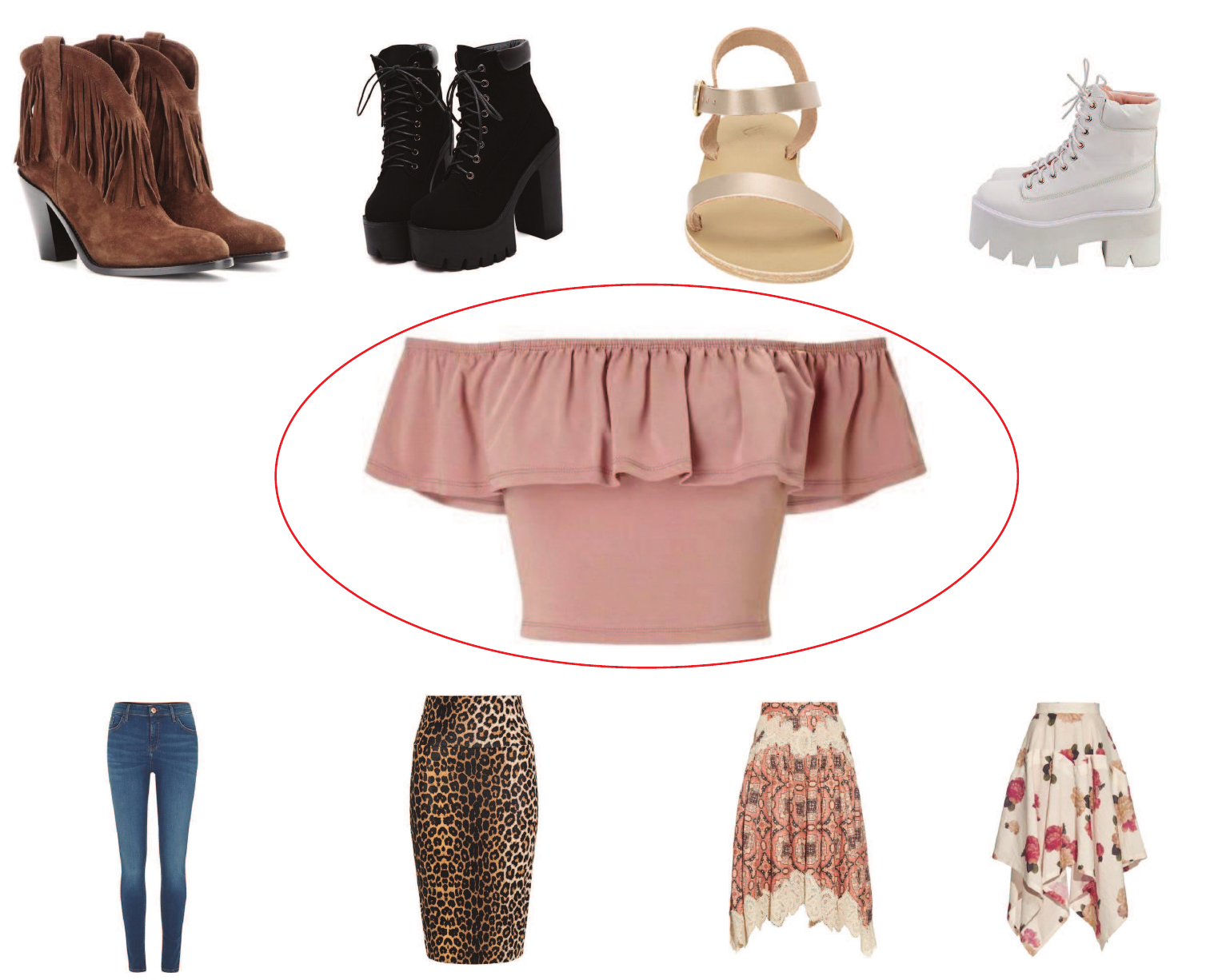} &
     \includegraphics[width=0.11\textwidth]{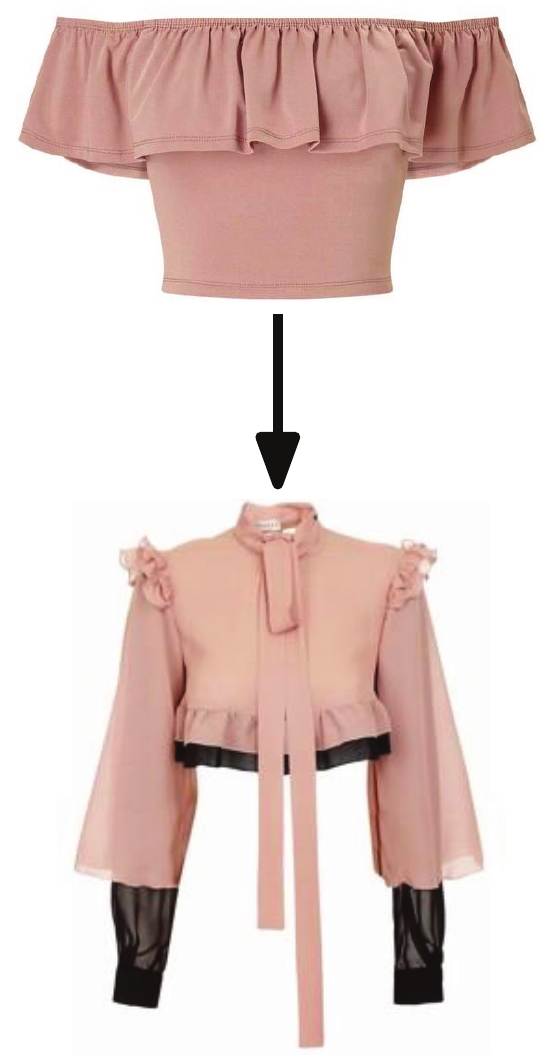} \\
    \end{tabular}
  \caption{\label{fig.diversity} Notion of Compatibility (Left) \vs Similarity (Right). \textbf{Left:} The upper outer garment in the center (red circle) is the query item. The surrounding items are its compatible ones. The styles of both the compatible shoes and lower body garments are various. \textbf{Right:} The style of a similar item (bottom) is constrained.}
\end{figure}

For items with a sufficient number of viewing or purchasing intents,
it is possible to take the co-viewing (or co-purchasing) records as signals of compatibility,
and simply use standard techniques for a recommendation system, such as collaborative filtering,
to identify compatible items.
In real world application, it is quite often encountered that there are insufficient records
to make a decent compatible recommendation --- it is then critical to fully exploit
relevant contents associated with items, such as the images for dresses, or the wineries for wines.

Even leveraging such relevant information, recommending or generating compatible items is
challenging due to three key reasons.
First, the notion of compatibility typically goes across categories and is broader
and more diverse than the notion of similarity, and it involves complex many-to-many relationships.
As shown in Figure~\ref{fig.diversity}, compatible items are not necessarily similar and vice versa.
Second, the compatibility relationship is inherently asymmetric for real world applications.
For instance, students purchase elementary textbooks before buying advanced ones, house owners buy furniture
only after their house purchases. Recommendation systems must take the asymmetry into consideration,
as recommending car accessories to customers who bought cars is rational;
recommending cars to those who bought car accessories would be improper.
The two reasons above make
many existing methods~\cite{McATarShiHen15,VeiKovBelMcABalBel15,2017arXiv170804014L} less fit for compatibility
learning, as they aim to learn a symmetric metric so as to model the item-item relationship.
Third, the currently available labeled data sets of compatible and incompatible items are insufficient
to train a decent image generation model.
Due to the asymmetric relationships, the generator could not simply learn to modify the input image as most CGANs do in the similarity learning setting.

However, humans have the capabilities to create compatible items by associating internal concepts.
For instance, fashion designers utilize their internal concept of compatibility, e.g., style and material to design many compatible outfits.
Inspired by this, we demonstrate extracting meaningful representation from the image contents for compatibility
is an effective way of tackling such challenges.

We aim at recommending and generating compatible items through learning
a ``Compatibility Family''.
The family for each item contains a representation vector as the embedding of the item, and multiple compatible prototype
vectors in the same space.
We refer to the latent space as the ``Compatibility Space''.
Firstly, we propose an end-to-end trainable system to learn the family for each item.
The multiple prototypes in each family capture the diversity of compatibility, conquering the first challenge.
Secondly, we introduce a novel Projected Compatibility Distance (PCD) function which is differentiable
and ensures diversity by encouraging the following properties:
(1) at least one prototype is close to a compatible item, (2) none of the prototypes is close to an incompatible item.
The function captures the notion of asymmetry for compatibility, tackling the second challenge.
While our paper focuses mainly on image content, this framework can also be applied to other modalities.

The learned compatible family's usefulness is beyond item recommendation.
We design a compatible image generator, which can be trained with only the limited labeled data
given the succinct representation that has been captured in the compatibility space, bypassing the third challenge.
Instead of directly generating the image of a compatible item from a query item,
we first obtain a compatible prototype using our system.
Then, the prototype is used to generate images of compatible items.
This relieves the burden for the generator to simultaneously learn the notion of compatibility
and how to generate realistic images.
In contrast, existing approaches generate target images directly from source images or source-related features. We propose a novel generator referred to as Metric-regularized Conditional Generative Adversarial Network (MrCGAN).
The generator is restricted to work in a similar latent space to the compatibility space.
In addition, it learns to avoid generating ambiguous samples that lie
on the boundary of two clusters of samples that have conflicting relationships with some query items.

We evaluate our framework on Fashion-MNIST dataset, two Amazon product datasets,
and Polyvore outfit dataset. Our method consistently achieves state-of-the-art performance
for compatible item recommendation.
Finally, we show that we can generate images of compatible items using our learned Compatible Family and MrCGAN.
We ask human evaluators to judge the relative compatibility between our generated images
and images generated by CGANs conditioned directly on query items.
Our generated images are roughly 2x more likely to be voted as compatible.

The main contributions of this paper can be summarized as follows:
\begin{itemize}
\item Introduce an end-to-end trainable system for Compatible Family learning to capture asymmetric-relationships.
\item Introduce a novel Projected Compatibility Distance to measure compatibility given limited ground truth compatible and incompatible items.
\item Propose a Metric-regularized Conditional Generative Adversarial Network model to visually reveal our learned compatible prototypes.
\end{itemize}

\section{Related Work}\label{sec.rw}
We focus on describing the related work in content-based compatible item recommendation
and conditional image generation using Generative Adversarial Networks (GANs).

\subsubsection{Content-based compatible item recommendation}

Many works assume a similarity learning setting that requires compatible items to stay close in a learned latent space.
\citeauthor{McATarShiHen15}~\shortcite{McATarShiHen15} proposed to use Low-rank
Mahalanobis Transform to map compatible items to embeddings close in the latent space.
\citeauthor{VeiKovBelMcABalBel15}~\shortcite{VeiKovBelMcABalBel15} utilized the co-purchase records from Amazon.com
to train a Siamese network to learn representations of items.
\citeauthor{2017arXiv170804014L}~\shortcite{2017arXiv170804014L} assumed that different items in
an outfit share a coherent style, and proposed to learn style representations of fashion items
by maximizing the probability of item co-occurrences.
In contrast, our method is designed to learn asymmetric-relationships.

Several methods go beyond similarity learning.
\citeauthor{IJCAI112887}~\shortcite{IJCAI112887} proposed to learn a topic model to find compatible tops from bottoms.
\citeauthor{HePacMcA16}~\shortcite{HePacMcA16} extended the work of \citeauthor{McATarShiHen15}~\shortcite{McATarShiHen15} to compute ``query-item-only" dependent weighted sum (i.e., independent of the compatible items) of $ \K $ distances
between two items in $ \K $ latent spaces to handle heterogeneous item recommendation.
However, this means that the query item only prefers several subspaces.
While this could deal with diversity across different query items (i.e., different query items have different compatible items), it is less effective for diversity across compatible items of the same query item (i.e., one query item has a diverse set of compatible items).
Our model instead represents the distance between an item and a candidate as the minimum of the distance between each prototype and the candidate, allowing compatible items to locate on different locations in the same latent space.
In addition, our method is end-to-end trained, and can be coupled with MrCGAN to generate images of compatible items,
unlike the other methods.

Another line of research tackles outfit composition as a set problem, and attempts to predict the most compatible item in a multi-item set.
\citeauthor{Li2017MiningFO}~\shortcite{Li2017MiningFO} proposed learning the representation of an outfit by pooling the item features produced by a multi-modal fusion model. Prediction is made by computing a compatibility score for the outfit representation.
\citeauthor{han2017learning}~\shortcite{han2017learning} treated items in an outfit as a sequence and modeled the item recommendation with a bidirectional LSTM to predict the next item from current ones. Note that these methods require mult-item sets to be given. This potentially can be a limitation in practical use.

\subsubsection{Generative Adversarial Networks}
After the introduction of Generative Adversarial Networks (GANs) \cite{NIPS2014_5423} as a popular way to train generative models,
GANs have been shown to be capable of conditional generation \cite{DBLP:journals/corr/MirzaO14}, which
has wide applications such as
image generation by class labels \cite{pmlr-v70-odena17a}
or by texts \cite{reed2016generative}, and image transformation between different
domains \cite{pix2pix2016,pmlr-v70-kim17a,CycleGAN2017}.
Most works for conditional generation focused on similarity relationships.
However, compatibility is represented by complex many-to-many relationships,
such that the compatible item could be visually distinct from the query item and
different query items could have overlapping compatible items.

The idea to regularize GANs with a metric is related to the reconstruction loss
used in autoencoders that requires the reconstructed image to stay close to
the original image, and it was also applied to GANs by
comparing the visual features extracted from samples to enforce
the metric in the sample space \cite{2015arXiv151209300B,2016arXiv161202136C}.
Our model instead regularizes the subspace of the
latent space (\ie, the input of the generator).
The subspace does not have the restriction to have known distribution
that could be sampled, and
visually distinct samples are allowed to be close as
long as they have similar compatibility relationships with other items.
This allows the subspace to be learned from a far more powerful
architecture.

\section{Our Method}\label{sec.method}
We first introduce a novel Projected Compatibility Distance (PCD) function.
Then, we introduce our model architecture and learning objectives.
Finally, we introduce a novel metric-regularized Conditional GAN (MrCGAN).
The notation used in this paper is also summarized in Table~\ref{tbl.notation}.

\begin{table}
\centering
\begin{tabular}{ll}
                 & Explanation                       \\ \hline
$ \K  $          & the number of prototypes \\
$ \N  $          & the size of the latent vector \\
$  X  $          & the set of query items \\
$  Y  $          & the set of items to be recommended \\
$ E_0(\cdot)  $  & the encoding function \\
$ \Ek{0}{y} $    & a shorthand for $ E_0(y) $ \\
$ E_k(\cdot) $   & the k-th projection, where $k\in\{1,\dots,\K\}$ \\
$ \Ek{k}{x} $    & a shorthand for $ E_k(x) $ \\
$ d(x, y) $      & PCD between $x$ and $y$\\
$ d_k(x, y) $    & squared $L_2$ distance between $ \Ek{k}{x} $ and $ \Ek{0}{y} $\\
\hline
$ G   $          & the generator \\
$ D   $          & the discriminator \\
$ Q_0(\cdot) $   & the latent vector prediction from $ D $ \\
$ \Vk{q}{0}{y}$  & a shorthand for $ Q_0(y) $ \\
$ z   $          & the noise input to the generator  \\
$ p_z $          & the distribution of $ z $\\
\hline
\end{tabular}
\caption{\label{tbl.notation} Notation. }
\end{table}

\subsection{Projected Compatibility Distance}
PCD is proposed to measure
the compatibility between two items $ x $ and $ y $.
Each item is transformed into a latent vector by
an encoding function $ E_0(\cdot) $. Additionally, $ \K $ projections, denoted as $ \{E_k(\cdot)\}_{k\in{\{1,\dots,\K\}}} $, are learned
to directly map an item to $ \K $ latent vectors (\ie, prototypes) close
to clusters of its compatible items. Each of the latent vectors has a size of $ \N $.
Finally the compatibility distance is computed as follows,

\begin{equation}
d(x, y) = \left\|
\frac {\sum_{k=1}^{\K}{ [\exp(-d_k(x,y))\Ek{k}{x}}] }
      {\sum_{k=1}^{\K}{ \exp(-d_k(x,y))} }
- \Ek{0}{y}
\right\|_2^2
\label{eq.pcd}
\end{equation}
where
\begin{equation}
d_k(x,y) = \left\| \Ek{k}{x}-\Ek{0}{y} \right\|_2^2,
\end{equation}
and $ \Ek{k}{x} $ stands for $ E_k(x) $ for readability.

\begin{figure}
  \centering
    \includegraphics[width=0.38\textwidth]{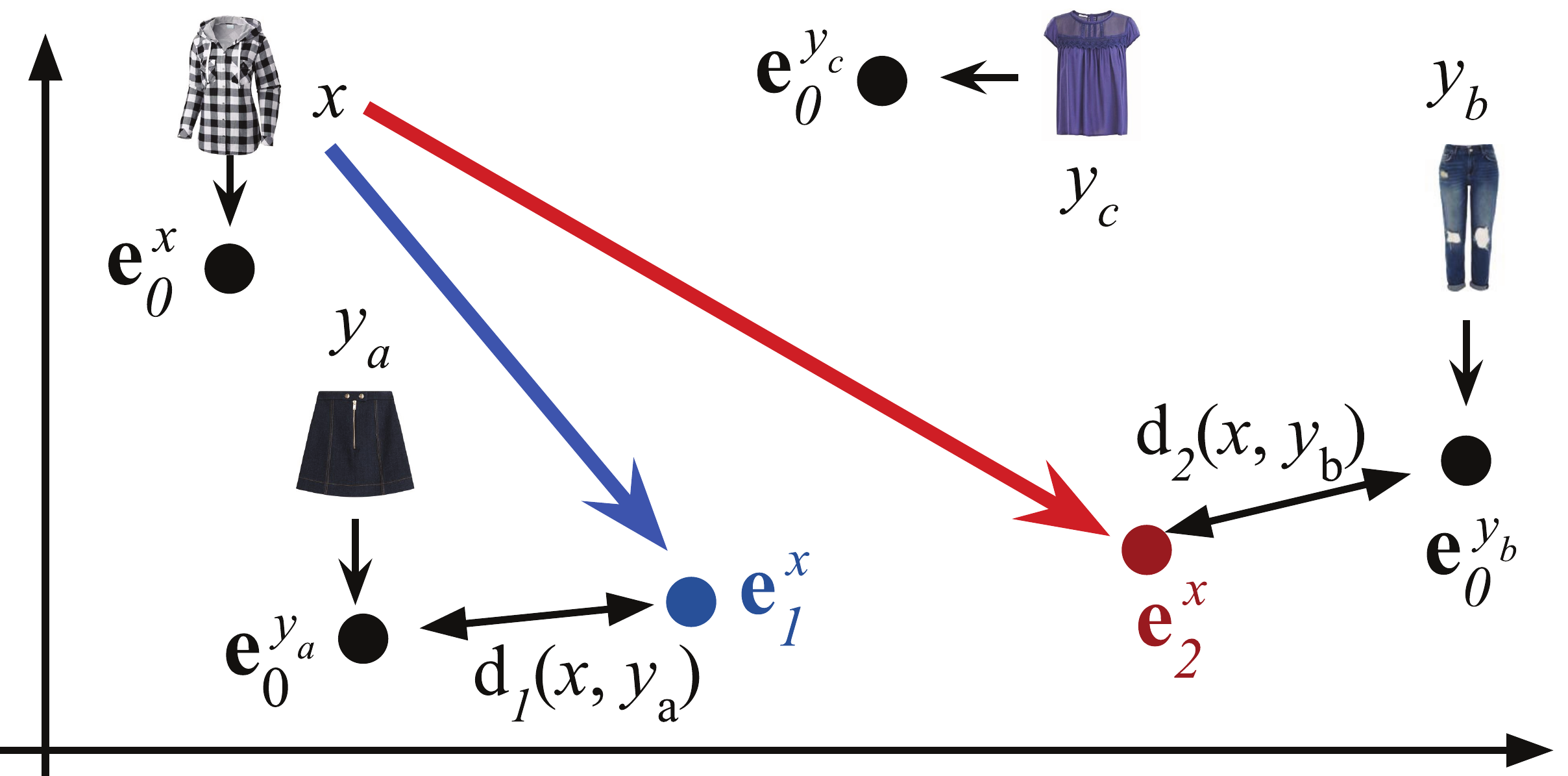}
  \caption{\label{fig.distance} Projected Compatibility Distance. A query item $x$ is respectively projected to $ \Ek{1}{x} $ and $ \Ek{2}{x} $ for its two distinct compatible items, $ y_a $ and $ y_b $. Thus, $d(x,y_a)\sim d_1(x,y_a) $ for $y_a$ while $d(x,y_b)\sim d_2(x,y_b) $ for $y_b$. For an incompatible item $y_c$, none of the projections are close to $\Ek{0}{y_c}$.
  }
\end{figure}

The concept of PCD is illustrated in Figure~\ref{fig.distance}.
When at least one $ \Ek{k}{x} $ is close enough to a latent vector $ \Ek{0}{y} $ of item $ y $,
the distance $ d(x, y) $ approaches $ \min_k d_k(x, y) $.

\subsection{Model Architecture}

\begin{figure}
  \centering
    \includegraphics[width=0.47\textwidth]{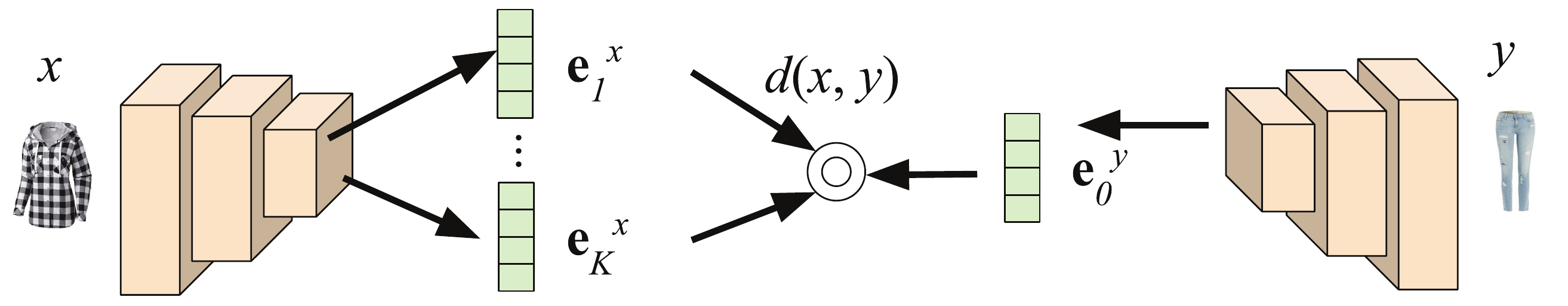}
  \caption{\label{fig.model} Model Architecture. The CNNs on the left and right are identical. Only the prototypes $\{\Ek{k}{x}\}_{k\in\{1,\dots,K\}}$ from $x$ and $\Ek{0}{y}$ from $y$ are considered to form (\ref{eq.pcd}).}
\end{figure}

In our experiments, most of the layers between $ E_k(\cdot) $ are shared and only the last layers for outputs are separated.
This is achieved by using Siamese CNNs \cite{Hadsell06dimensionalityreduction} for feature transformation. As illustrated in Figure~\ref{fig.model}, rather than learning just an embedding like the original formulation, $ \K + 1 $ projections are learned for the item embedding $ E_0(\cdot) $ and $ \K $ prototype projections.

We model the probability of a pair of items $(x, y)$ being compatible with a
shifted-sigmoid function similar to \cite{HePacMcA16} as shown below,
\begin{equation}
P(x, y) = \sigma_c(-d(x, y)) = \frac{1}{1+ \exp(d(x, y) - c)}~,
\end{equation}
where $c$ is the shift parameter to be learned.

\subsubsection{Learning objective}
Given the set of compatible pairs $R^+$ and the set of incompatible pairs $R^-$,
we compute the binary cross-entropy loss, \ie,
\begin{align}
\begin{split}
L_{ce} &= - \frac{1}{|R^+|}\sum_{(x, y) \in R^+} \log \left(P(x, y) \right) \\
  &- \frac{1}{|R^-|}\sum_{(x, y) \in R^-}\log \left(1 - P(x, y)\right)~,
\end{split}
\end{align}
where $|\cdot|$ denotes the size of a set.
We further regularize the compatibility space by minimizing the distance for compatible pairs. The total loss is as follows:
\begin{equation}
L = L_{ce} + \lambda_{m} \frac{1}{|R^+|}\sum_{(x,y)\in R^+} d(x, y)~,
\end{equation}
where $\lambda_{m}$ is a hyper-parameter to balance loss terms.

\subsection{MrCGAN for Compatible Item Generation}
\begin{figure}
  \centering
    \includegraphics[width=0.46\textwidth]{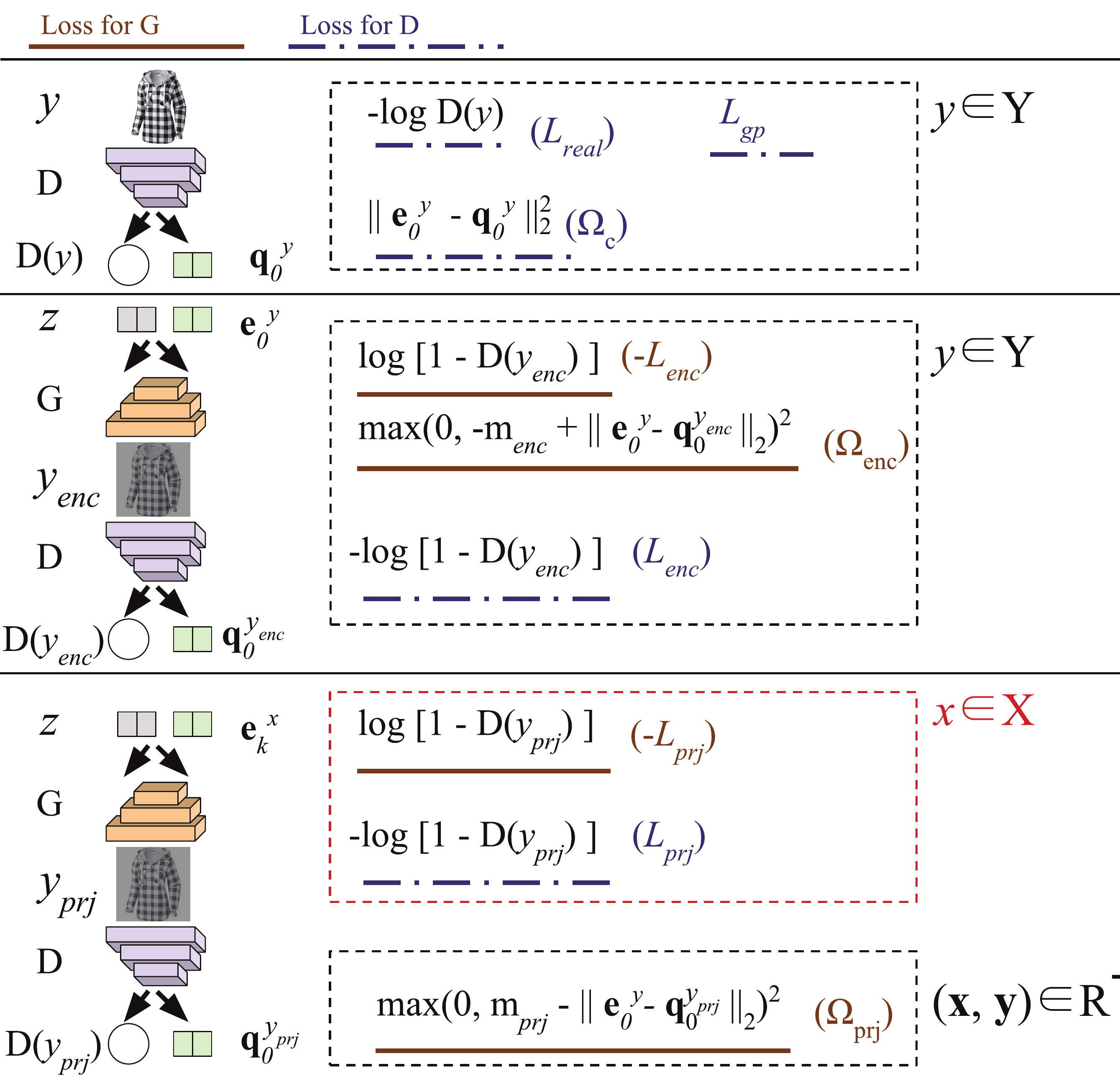}
  \caption{\label{fig.mrcgan} The training procedure of MrCGAN.
  The generated samples conditioned on different latent vectors are denoted as:
  $ y_{enc} = G(z, \Ek{0}{y})$, and $ y_{prj} = G(z, \Ek{k}{x}) $.}
\end{figure}

Once the compatibility space is learned,
the metric within the space could be used to regularize a CGAN.
The proposed model is called Metric-regularized CGAN (MrCGAN).
The learning objectives are illustrated in Figure~\ref{fig.mrcgan}.

The discriminator has two outputs: $ D(y) $, the probability of the sample $ y $ being real,
and $ Q_0(y) = \Vk{q}{0}{y} $, the predicted latent vector of $ y $.
The generator is conditioned
on both $ z $ and a latent vector from the compatibility space
constructed by $ E_0(\cdot)$. Given the set of query items $ X $, items to be recommended $ Y $, and
the noise input $ z \in \mathbb{R}^Z \sim p_z = \mathcal{N}(0, 1) $,
we compute the MrCGAN losses as,
\begin{align}
L_{real} &= -\frac{1}{|Y|}\sum_{y \in Y} \log {D(y)}~, \\
L_{enc} &= -\frac{1}{|Y|}\sum_{y \in Y}\mathop{\mathbb{E}}_{ z \sim p_z }
  \log { (1-D(G(z, \Ek{0}{y}))) }~, \\
L_{prj} &= -\frac{1}{|\K| | X |}\sum_{k=1}^{\K}\sum_{x \in X}
  \mathop{\mathbb{E}}_{ z \sim p_z }
  \log { (1-D(G(z, \Ek{k}{x}))) }
\end{align}
The discriminator learns to discriminate
between real and generated images,
while the generator learns to fool the discriminator with both
the encoded vector $ \Ek{0}{y} $ and the projected prototypes $ \Ek{k}{x} $ as
conditioned vectors.
We also adopt the gradient penalty loss $ L_{gp} $
of DRAGAN \cite{2017arXiv170507215K},
\begin{equation}
L_{gp} = \lambda_{gp}
	\mathop{\mathbb{E}}_{\hat{y}\sim p_{perturbed~Y}}
    (\left\|\nabla_{\hat{y}} D(\hat{y}) \right\|_2 - 1) ^2~,
\end{equation}
where a perturbed batch is computed from a batch sampled from $ Y $ as
$ batch + \lambda_{dra} \cdot batch.stddev() \cdot \mathcal{U}[0, 1] $,
and $ \lambda_{gp} $, $ \lambda_{dra} $ are hyper-parameters.
In addition, MrCGAN has the following metric regularizers,
\begin{equation}
\Omega_c = \frac{1}{|Y|}\sum_{y \in Y} \left\| \Ek{0}{y} - \Vk{q}{0}{y} \right\|_2^2~,
\end{equation}
which requires the predicted $ \Vk{q}{0}{y} $ to stay close to
the real latent vector $ \Ek{0}{y} $, forcing $ Q_0(\cdot) $ to approximate $ E_0(\cdot) $,
and
\begin{equation}
\Omega_{enc} = \frac{1}{|Y|}\sum_{y \in Y} M^+(\Ek{0}{y}, \Ek{0}{y})~,
\end{equation}
where $ m_{enc} $ is a hyper-parameter and
\begin{equation*}
M^+(v, s) = \mathop{\mathbb{E}}_{z\sim p_z}
	\max(0, - m_{enc} + \left\| v - Q_0(G(z, s)) \right\|_2)^2~,
\end{equation*}
which measures the distance between a given vector
$ v $ and the predicted latent vector of the generated sample conditioned on $ s $, and
it guides the generator to learn to align its latent space with the compatibility space.
The margin $ m_{enc} $ relaxes the constraint so that the generator does not
collapse into a 1-to-1 decoder.
Finally the generator also learns to avoid generating incompatible items by,
\begin{align}
\Omega_{prj} &= \frac{1}{|\K||R^-|}\sum_{k = 1}^{\K} \sum_{(x, y) \in R^-}
  M^-(\Ek{0}{y}, \Ek{k}{x})~,
\end{align}
where $ m_{prj} $ is a hyper-parameter and

\begin{equation*}
M^-(v, s) = \mathop{\mathbb{E}}_{z\sim p_z}
	\max(0, m_{prj} - \left\| v - Q_0(G(z, s)) \right\|_2)^2~,
\end{equation*}
which requires the distance between a given latent vector $ v $ and
the predicted latent vector of the generated sample conditioned on $ s $ to be larger
than a margin $ m_{prj} $.

The total losses for $ G $ and $ D $ are as below,
\begin{align}
L_G &= -\frac{1}{2}(L_{enc} + L_{prj}) + \Omega_{enc} + \Omega_{prj} \\
L_D &= L_{real} + \frac{1}{2}(L_{enc} + L_{prj}) + L_{gp} + \Omega_{c}
\end{align}

In effect, the learned latent space is constructed by:
(1) $ z $ space, (2) a subspace that has similar structure as the compatibility space.
To generate compatible items for $ x $, $ \Ek{k}{x} $ is used as the
conditioning vector: $ G(z, \Ek{k}{x}) $.
To generate items with similar style
to $ y $, $ \Ek{0}{y} $ is used instead: $ G(z, \Ek{0}{y}) $.

\subsection{Implementation Details}
We set $ \lambda_{m} $ to 0 and 0.5 respectively for recommendation and generation experiments and the batch size to 100, and use Adam optimizer with $(\lambda_{lr}, \beta_1, \beta_2) = (0.001, 0.9, 0.999)$. The validation set is for the best epoch selection.
The last layer of each discriminative model takes a fully-connected layer with weight normalization \cite{NIPS2016_6114} except for the Amazon also-bought/viewed experiments, where the weight normalization is not used for fair comparison.
Before the last layer, the following feature extractors are for different experiments:
(1) Fashion-MNIST+1+2 / MNIST+1+2: multi-layer CNNs with weight normalization,
(2) Amazon also-bought/viewed: none, (3) Amazon co-purchase: Inception-V1 \cite{43022}, (4) Polyvore:
Inception-V3 \cite{Szegedy_2016_CVPR}.

\begin{figure*}
  \centering
    \includegraphics[width=0.80\textwidth]{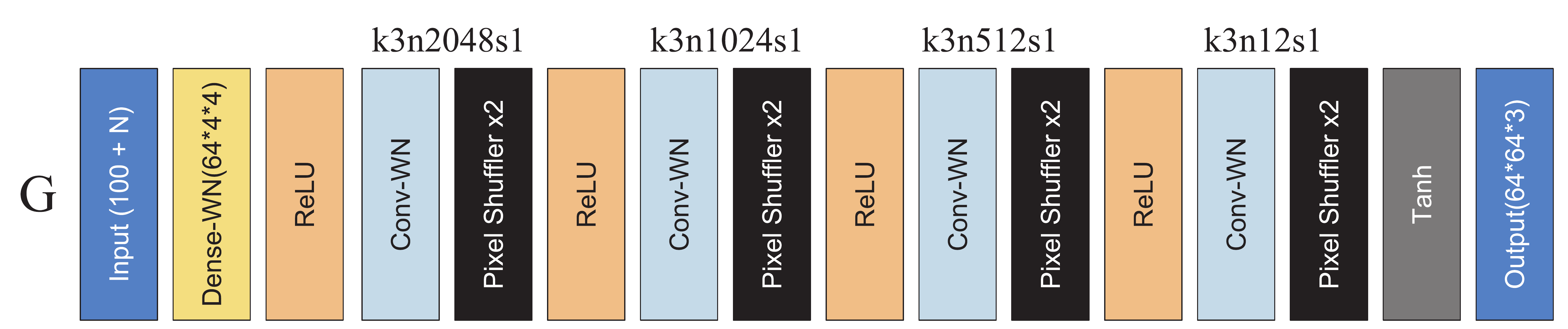}
    \includegraphics[width=0.98\textwidth]{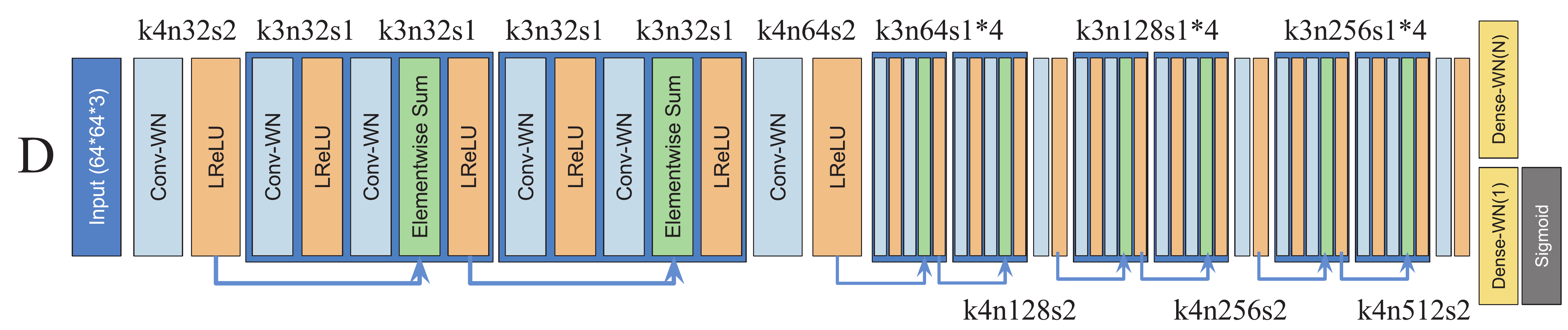}
  \caption{\label{fig.srgan} MrCGAN Architecture}
\end{figure*}

For the generation experiments,
we set $ \lambda_{gp}$ to 0 and apply DCGAN \cite{2015arXiv151106434R} architecture for both of our model and the GAN-INT-CLS \cite{reed2016generative} in MNIST+1+2. For Amazon
co-purchase and Polyvore generation experiments, we set both $ \lambda_{gp}$ and $\lambda_{dra}$ to 0.5, and adopt a SRResNet\cite{2016arXiv160904802L}-like architecture for GAN training with a different learning rate setting, \ie, $ (\lambda_{lr}, \beta_1, \beta_2) = (0.0002, 0.5, 0.999) $.
The model and parameter choosing is inspired by \citeauthor{2017arXiv170805509J} \shortcite{2017arXiv170805509J}, but we take off the skip connections from the generator since it does not work well in our experiment, and we also use weight normalization for all layers. The architecture is shown in Figure~\ref{fig.srgan}.

In most of our experiments, the sets, $ X $ and $ Y $, are identical. However, we create non-overlapped sets of $ X $ and $ Y $ by restricting the categories in the generation experiments for Amazon co-purchase and Polyvore dataset. The dimension of $z$ and the number of $\K$ are respectively set to 20 and 2 in all generation experiments.
Besides, the rest of the parameters are taken as follows:
(1) MNIST+1+2: $ (\N, m_{enc}, m_{prj})$ = $(20, 0.1, 0.5)$,
(2) Amazon co-purchase:  $ (\N, m_{enc}, m_{prj})$ = $(64, 0.05, 0.2)$,
(3) Polyvore: $ (\N, m_{enc}, m_{prj})$ = $(20, 0.05, 0.3)$.

\section{Experiment}\label{sec.exp}

We conduct experiments on several datasets and compare the performance with state-of-the-art methods
for both compatible item recommendation and generation.

\subsection{Recommendation Experiments}

\subsubsection{Baseline} Our proposed PCD is compared with two baselines:
(1) the $L_2$ distance between the latent vectors of the Siamese model,
(2) the Monomer proposed by \citeauthor{HePacMcA16} \shortcite{HePacMcA16}.
Although Monomer was originally not trained
end-to-end, we still cast an end-to-end setting for it on Fashion-MNIST+1+2 dataset.

Our model achieves superior performance in most experiments. In addition, our model has two advantages in efficiency compared to Monomer:
(1) our storage is $ \frac{1}{\K} $ of Monomer
since they projected each item into $ \K $ spaces
beforehand while we only compute $ \K $ prototype projections during
recommendation,
(2) PCD is approximately $ \min_k d_k(x, y)$. Therefore, in query time we could do $ \K $ nearest-neighbor search in parallel to get approximate results, while Monomer needs to aggregate the weighted sum of the distances in $ \K $ latent spaces.

\subsubsection{Fashion-MNIST+1+2 Dataset}

\begin{table}
\centering
\begin{tabular}{|c|c|c|c|c|}
\hline

Model                      &  $\K$  & Error rate  & AUC       \\ \hline
$L_2$                    &     & 41.45\% $ \pm $ 0.55\%  &  0.6178 $ \pm $ 0.0034 \\ \hline
\multirow{4}*{Monomer} &  2  & 23.24\% $ \pm $ 0.62\%  &  0.8533 $ \pm $ 0.0045 \\
                           &  3  & 22.46\% $ \pm $ 0.44\%  &  0.8598 $ \pm $ 0.0021 \\
                           &  4  & 21.60\% $ \pm $ 0.39\%  &  0.8651 $ \pm $ 0.0052 \\
                           &  5  & 22.03\% $ \pm $ 0.39\%  &  0.8623 $ \pm $ 0.0030 \\\hline
\multirow{4}*{PCD}    &  2  & 21.31\% $ \pm $ 0.77\%  &  0.8746 $ \pm $ 0.0070 \\
                           &  3  & 20.66\% $ \pm $ 0.30\%  &  0.8804 $ \pm $ 0.0033 \\
                           &  4  & 20.39\% $ \pm $ 0.60\%  &  0.8808 $ \pm $ 0.0043 \\
                           &  5  & \textbf{20.01\%} $ \pm $ 0.39\%  &  \textbf{0.8830} $ \pm $ 0.0032 \\ \hline
\end{tabular}
\caption{\label{tbl.fashion} Performance on Fashion-MNIST+1+2. }
\end{table}

To show our model's ability to handle asymmetric relationships, we build a toy dataset from Fashion-MNIST \cite{xiao2017/online}. The dataset consists of 28x28 gray-scale images of 10 classes, including T-shirt, Trouser, Pullover, etc.
We create an arbitrary asymmetric relationship as follows,
\begin{equation*}
 (x, y) \in R^+ \iff C_y = (C_x + i) \mod 10, \forall i\in\{1,2\},
\end{equation*}
where $C_x$ means the class of $x$. The other cases of $ (x, y)$ belong to $R^- $.

Among the 60,000 training samples, 16,500 and 1,500 pairs are non-overlapped and randomly selected to form the training and validation sets, respectively. Besides, 10,000 pairs are created from the 10,000 testing samples for the testing set. The samples in each split are also non-overlapped. The strategy of forming a pair is that we randomly choose a negative or a positive sample $y$ for each sample $x$. Thus, $|R^+|\approx|R^-|$. We erase the class labels while only keep the pair labels during training for the reason that the underlying factor for compatibility is generally unavailable.

We repeat the above setting five times and show the averaged result in Table~\ref{tbl.fashion}. Like the settings of \citeauthor{HePacMcA16}~\shortcite{HePacMcA16}, the latent size in $L_2$ equals to $ (\K+1)\times \N$. Here, the size in $L_2$ is 60. Each model is trained for 50 epochs. The experiment shows  that the $L_2$ model performs poorly on a highly asymmetric dataset while our model achieves the best results.

\subsubsection{Amazon also-bought/viewed Dataset}
The image features of 4096 dimensions are extracted beforehand in this dataset. Following the setting of \citeauthor{HePacMcA16}~\shortcite{HePacMcA16}, the also-bought and also-viewed relationships in the Amazon dataset \cite{McATarShiHen15} are positive pairs while the negative pairs are sampled accordingly. We set our parameters, $\K$ and $\N$, as the same as Monomer for comparison, \ie, (1) $ (\K, \N) = (4, 20) $ for also-bought, and (2) $(\K, \N)=(4, 10) $ for also-viewed. We train 200 epochs for each model and Table~\ref{tbl.monomer} shows the results. Compared to the error rates reported in \citeauthor{HePacMcA16}~\shortcite{HePacMcA16}, our model yields the best performance in most settings and the lowest error rate on average.

\begin{table}
\centering
\begin{tabular}{|c c c c c|}
\hline
Dataset & Graph & LMT & Monomer & PCD \\ [0.5ex]
\hline
\multirow{2}*{Men} & \multirow{2}{5em}{also\_bought also\_viewed} & 9.20\% & 6.48\% & \textbf{6.05\%}\\
& & 6.78\% & 6.58\% & \textbf{5.97\%} \\
\hline
\multirow{2}*{Women}  & \multirow{2}{5em}{also\_bought also\_viewed} & 11.52\% & 7.87\% & \textbf{7.75\%} \\
& & 7.90\% & \textbf{7.34\%} & 7.37\% \\
\hline
\multirow{2}*{Boys}  & \multirow{2}{5em}{also\_bought also\_viewed} & 8.80\% & 5.71\% & \textbf{5.27\%} \\
& & 6.72\% & 5.35\% & \textbf{5.03\% }\\
\hline
\multirow{2}*{Girls}  & \multirow{2}{5em}{also\_bought also\_viewed} & 8.33\% & 5.78\% & \textbf{5.34\%} \\
& & 6.46\% & 5.62\% & \textbf{4.86\%} \\
\hline
\multirow{2}*{Baby}  & \multirow{2}{5em}{also\_bought also\_viewed} & 12.48\% & 7.94\% & \textbf{7.00\%} \\
& & 11.88\% & 9.25\% & \textbf{8.00\%} \\ \hline
Avg. & &  9.00\% & 6.79\%  & \textbf{6.26\%} \\ \hline
\end{tabular}
\caption{\label{tbl.monomer} Error rates on Amazon also-bought/viewed dataset. LMT stands for Low-rank Mahalanobis Transform \cite{McATarShiHen15}.}
\end{table}

\subsubsection{Amazon Co-purchase Dataset}

Based on the data split\footnote{https://vision.cornell.edu/se3/projects/clothing-style/}$^,$\footnote{7 images from the training set are missing, so 35 pairs containing these images are dropped.} defined in \citeauthor{VeiKovBelMcABalBel15}~\shortcite{VeiKovBelMcABalBel15}, we increase the validation set via randomly selecting an additional 9,996 pairs from the original training set since its original size is too small, and accordingly decrease the training set by removing the related pairs for the non-overlapping requirement. Totally, 1,824,974 pairs remain in the training set.
As the ratio of positive and negative pairs is disrupted, we re-weigh each sample during training to re-balance it back to 1:16. Besides, we randomly take one direction for each pair since the ``Co-purchase'' relation in this dataset is symmetric.

We adopt and fix the pre-trained weights from \citeauthor{VeiKovBelMcABalBel15}~\shortcite{VeiKovBelMcABalBel15},
and replace the last embedding layer with each model's projection layer. The last layer is trained for 5 epochs in each model. Still, we set the latent size in \citeauthor{VeiKovBelMcABalBel15}~\shortcite{VeiKovBelMcABalBel15} as 256 and equal to $(\K+1)\times \N$. Thus, $(\K,\N)=(3, 64)$ for both Monomer and our model. Table~\ref{tbl.dyadic} shows the results. Our model obtains superior performance to \citeauthor{VeiKovBelMcABalBel15}~\shortcite{VeiKovBelMcABalBel15} and comparable results with Monomer.

\begin{table}
\centering
\begin{tabular}{|l|c|c|c|}
\hline

Model           &     AUC       \\ \hline
\citeauthor{VeiKovBelMcABalBel15} \shortcite{VeiKovBelMcABalBel15}        &     0.826     \\ \hline
\citeauthor{VeiKovBelMcABalBel15} \shortcite{VeiKovBelMcABalBel15} (retrain last layer) &    0.8698   \\ \hline
Monomer         &     0.8747 \\ \hline
PCD            &     \textbf{0.8762}    \\ \hline
\end{tabular}
\caption{\label{tbl.dyadic} AUC on Amazon co-purchase dataset. }
\end{table}

\subsubsection{Polyvore Dataset}
To demonstrate the ability of our model to capture the implicit nature of compatibility, we create an outfit dataset from Polyvore.com, a collection of user-generated fashion outfits. We crawl outfits under Women's Fashion category and group the category of items into tops, bottoms, and shoes. Outfits are ranked by the number of likes and the top 20\% are chosen as positive.

Three datasets are constructed for different recommendation settings:
(1) from tops to others, (2) from bottoms to others, and (3) from shoes to others. The statistics of the dataset are shown in Table~\ref{tbl.polyvore_data}.
The construction procedure is as follows:
(1) Items of source and target categories are non-overlapped split according to the ratios 60 : 20 : 20 for training, validation, and test sets.
(2) A positive pair is decided if it belongs to the positive outfit.
(3) The other combinations are negative and sub-sampled by choosing 4 items from target categories for each positive pair. Duplicate pairs are dropped afterwards.
This dataset is more difficult since the compatibility information across categories no longer exists, \ie, both positive and negative pairs are from the same categories. The model is forced to learn the elusive compatibility between items.

Pre-trained Inception-V3 is used to extract image features,
and the last layer is trained for 50 epochs.
We set $ \N$ to 100 for $L_2$, and $ (\K,\N) = (4,20)$ for Monomer and PCD.
The AUC scores are listed in Table~\ref{tbl.polyvore}, and our model still achieves the best results.

\begin{table}
\centering
 \begin{tabular}{|c c c c c|}
 \hline
 Dataset & split & \# source & \# target & \# pairs \\
 \hline
 \multirow{3}{6em}{top to others}     & train & 165,969 & 201,028 &  462,176  \\
                                      & val &   55,323  & 67,010  &  51,420   \\
                                      & test &  55,324  & 67,010  &  52,335   \\
 \hline
 \multirow{3}{6em}{bottom to others}  & train & 67,974 & 299,022 & 343,383 \\
                                      & val   & 22,659 & 99,675  & 40,015  \\
                                      & test  & 22,659 & 99,675  & 39,360  \\
 \hline

 \multirow{3}{6em}{shoe to others}    & train & 133,053 & 233,944 & 454,829 \\
                                      & val   & 44,351  & 77,982  & 48,500 \\
                                      & test  & 44,352  & 77,982  & 47,100 \\
 \hline
\end{tabular}
\caption{\label{tbl.polyvore_data} Polyvore dataset.}
\end{table}

\begin{table}
\centering
\begin{tabular}{|c c c c |}
\hline
Graph            & $L_2$& Monomer         & PCD             \\ [0.5ex]  \hline
top-to-others    & 0.7165 &         0.7431  & \textbf{0.7484} \\ \hline
shoe-to-others   & 0.6988 &         0.7146  & \textbf{0.7165} \\ \hline
bottom-to-others & 0.7450 &         0.7637  & \textbf{0.7680} \\ \hline
Avg.             & 0.7201 &         0.7405  & \textbf{0.7443} \\ \hline
\end{tabular}
\caption{\label{tbl.polyvore} AUC on Polyvore dataset.}
\end{table}

\subsection{Generation Experiments}
\subsubsection{Baseline} We compare our model with
GAN-INT-CLS from \cite{reed2016generative} in the MNIST+1+2 experiment.
$ \Ek{0}{x} $ is used as the conditioning vector for GAN-INT-CLS, and
each model uses exact architecture except that the discriminator for MrCGAN
outputs an additional $ Q_0(.) $, while the discriminator for GAN-INT-CLS is conditioned by $ \Ek{0}{x} $.
For other experiments, we compare with pix2pix \cite{pix2pix2016}
that utilizes labeled image pairs for image-to-image
translation and DiscoGAN \cite{pmlr-v70-kim17a} that
unsupervisedly learns to transform images into a different domain.

\begin{figure}[t]
  \centering
    \includegraphics[width=0.45\textwidth]{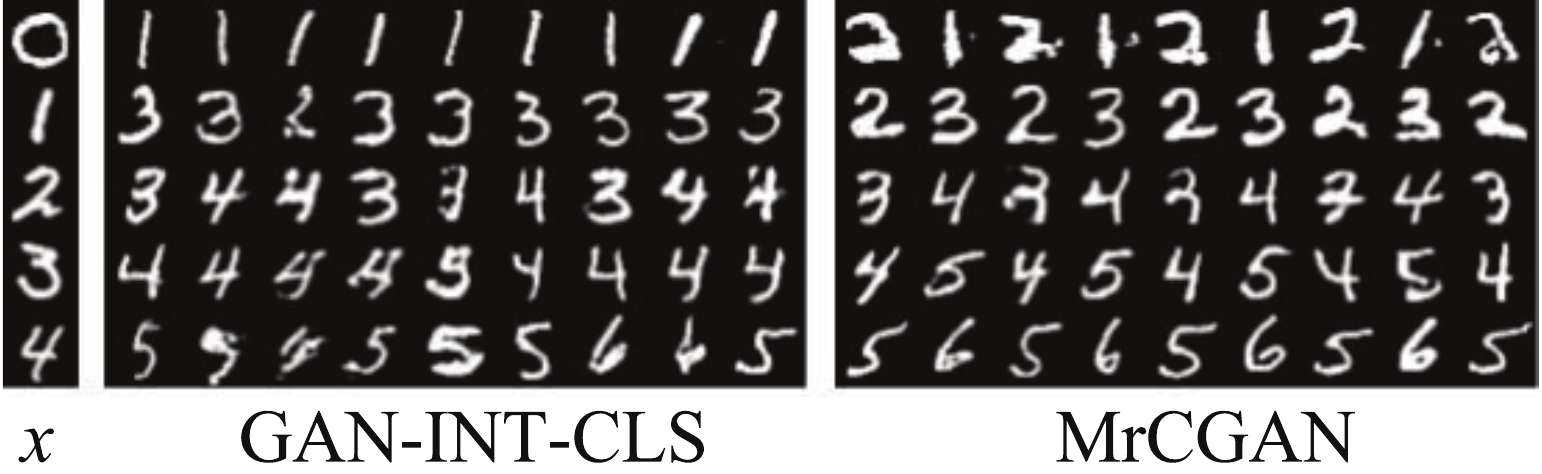} \\
  \caption{\label{fig.mnist} Column $x$ is the input. The rows show the generated images by varying $z$ and $k$ of different methods.
  }
\end{figure}

\begin{figure*}
  \centering
  \begin{subfigure}{\textwidth}
   \begin{tabular}{cc}
    Amazon Co-purchase & Polyvore Top-to-others \\
    \includegraphics[width=0.47\textwidth]{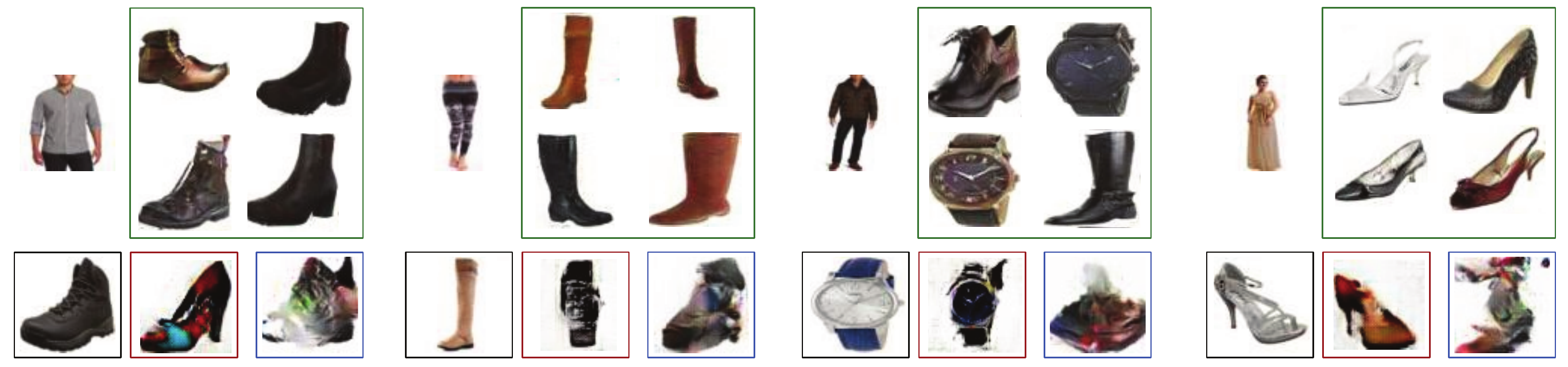} &
    \includegraphics[width=0.47\textwidth]{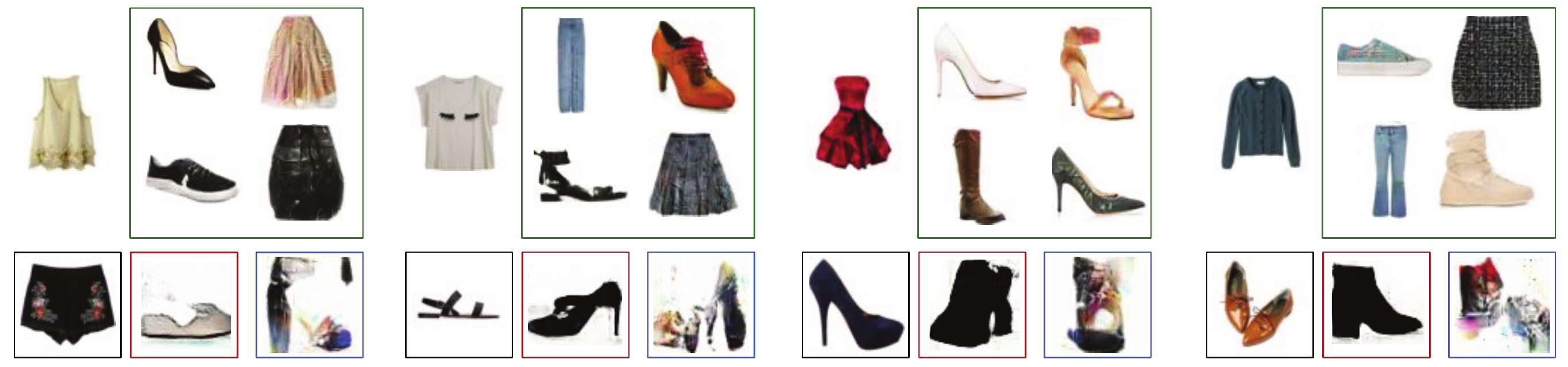} \\
   \end{tabular}
  \caption{\label{fig.project} Each block of images represents one set of conditional generation. \textbf{Top-left:} conditioning image $ x $. \textbf{Top-right: } four samples generated by MrCGAN conditioned on $ \Ek{k}{x} $ (in green box). \textbf{Bottom-left: } GT (in black box). \textbf{Bottom-middle: } DiscoGAN \cite{pmlr-v70-kim17a} (in red box). \textbf{Bottom-right:} pix2pix \cite{pix2pix2016} (in blue box).}
\end{subfigure}
\begin{subfigure}{\textwidth}
   \begin{tabular}{cc}
    \includegraphics[width=0.47\textwidth]{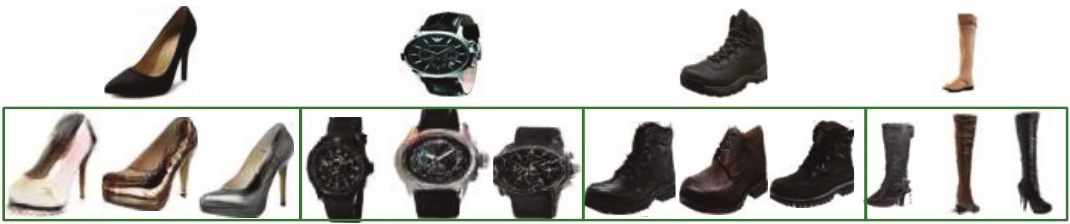} &
    \includegraphics[width=0.47\textwidth]{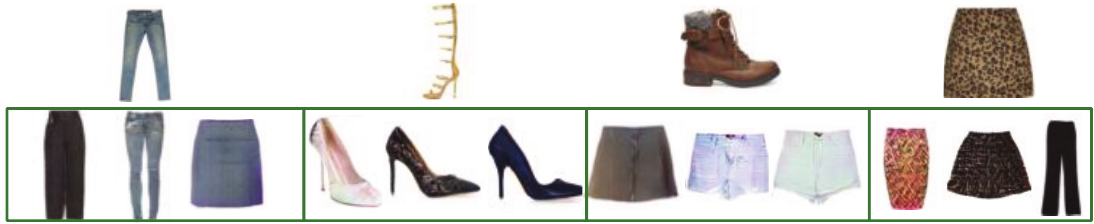} \\
   \end{tabular}
  \caption{\label{fig.near} Each block of images represents conditional generation (in green box) based on the latent vector $ \Ek{0}{y} $ of the image $ y $ on the top. Note that by conditioning on $ \Ek{0}{y} $, MrCGAN generates items having similar style as $ y $ instead of compatible items.}
\end{subfigure}
   \caption{\label{fig.mrcgan_compare}
   Examples of generated images conditioned on (a) $ \Ek{k}{x} $ and (b) $ \Ek{0}{y} $.}
\end{figure*}

\subsubsection{MNIST+1+2}
We use MNIST dataset to build a similar dataset as Fashion-MNIST+1+2
because the result is easier to interpret. Additional 38,500 samples are selected as unlabeled data
to train the GANs.
In Figure~\ref{fig.mnist} we display the generation results conditioned on samples
from the test set. We found that our model could preserve diversity better
and that the $ \K $ projections automatically group different modes,
so the variation can be controlled by changing $ k $.

\subsubsection{Amazon Co-purchase Dataset}
We sample a subset from Amazon Co-purchase
dataset \cite{VeiKovBelMcABalBel15}
by reducing the number of types for target items so that it's
easier for GANs to work with. In particular, we keep only relationships from Clothing to Watches and Shoes
in Women's and Men's categories.
We re-weigh each sample
during training to balance the ratio between positives and negatives to 1:1, but for validation set, we simply drop all excessive
negative pairs. Totally, there remain  9,176 : 86 : 557 positive pairs and
435,396 : 86 : 22,521 negative pairs for training, validation and
test split, respectively.
The unlabeled training set for GANs is selected from
the training ids from \citeauthor{VeiKovBelMcABalBel15}~\shortcite{VeiKovBelMcABalBel15}, and it consist of 226,566 and 252,476 items
of source and target categories, respectively. Each image is also resized to 64x64
before being fed into the discriminator.

Both DiscoGAN and pix2pix are inherently designed for similarity learning, and as
shown in Figure~\ref{fig.project}, they could not produce satisfying results.
Moreover, our model produces diverse outputs,
while the diversity for conventional image-to-image models are limited.
We additionally sample images conditioned on $ \Ek{0}{y} $ instead of on $ \Ek{k}{x} $
as illustrated in Figure~\ref{fig.near}.
We find that
the MrCGAN has the ability to generate diverse items having similar style as $ y $.

\subsubsection{Polyvore Top-to-others Dataset}
We use the top-to-others split as a demonstration
for our methods. Likewise we re-weigh each sample during training to
balance positive and negative pairs to 1:1 to encourage the model to
focus on positive pairs. The results are shown in Figure~\ref{fig.mrcgan_compare}.

\subsubsection{User study}

\begin{figure}
  \centering
    \begin{tikzpicture}
    \begin{axis}[
    ylabel=Total \# of Votes,
    enlarge x limits=0.2,
    legend style={
        at={(0.5,-0.25)},
        anchor=north,legend columns=-1
    },
    ymin=0,
    ybar,
    xtick=data,
    width=0.4\textwidth,
    height=0.25\textwidth,
    bar width=0.25cm,
    symbolic x coords={(a),(b),(c),(d)},
    xtick={(a),(b),(c),(d)},
    ymajorgrids=false,
    ]
    \addplot+[plot 0,bar group size={0}{3},error bars/.cd,y dir=both, y explicit]
    coordinates {
    ((a),589) +- (30.4150334005, 30.4150334005)
    ((b),625) +- (25.6647952248, 25.6647952248)
    };
    \addplot[plot 1,bar group size={1}{3},error bars/.cd,y dir=both, y explicit]
    coordinates {
    ((a),329) +- (29.1280834632, 29.1280834632)
    ((b),175) +- (23.186125045, 23.186125045)
    };
    \addplot[plot 2,bar group size={2}{3},error bars/.cd,y dir=both, y explicit]
    coordinates {
    ((a),76) +- (16.4486884739, 16.4486884739)
    ((b),60) +- (14.6718366237, 14.6718366237)
    };

    \addplot[plot 0,bar group size={0}{3},error bars/.cd,y dir=both, y explicit]
    coordinates {
    ((c), 348) +- (28.2116637886, 28.2116637886)
    ((d), 344) +- (28.936191144, 28.936191144)
    };
    \addplot[plot 3,bar group size={1}{3},error bars/.cd,y dir=both, y explicit]
    coordinates {
    ((c),222) +- (25.1775484351, 25.1775484351)
    ((d),328) +- (28.6364451655, 28.6364451655)
    };
    \addplot[plot 4,bar group size={2}{3},error bars/.cd,y dir=both, y explicit]
    coordinates {
    ((c),285) +- (27.0704176174, 27.0704176174)
    ((d),261) +- (26.9221280835, 26.9221280835)
    };

    \legend{MrCGAN,DiscoGAN,pix2pix,,Random,Decline}
    \end{axis}
    \end{tikzpicture}
  \caption{\label{fig.userstudy} Survey results on Amazon co-purchase (see panel (a, c)) and Polyvore (see panel (b, d)). MrCGAN (blue) generally outperforms others.}
\end{figure}
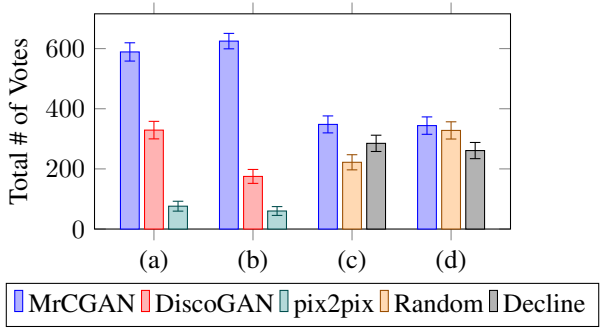

Finally we conduct online user surveys to see whether our model
could produce images that are perceived as compatible. We conduct two types of surveys:
(1) Users are given a random image from source categories and three generated images by different
models in a randomized order. Users are asked to select the item which is most compatible with the source item, and
if none of the items are compatible, select the most recognizable one.
(2) Users are given a random image from source categories, a random image from target categories, and
an image generated by MrCGAN in a randomized order. Users are asked to select the most compatible item with the source item, and if none
of the items are compatible, users can decline to answer.
The results are shown in Figure~\ref{fig.userstudy},
and it shows that MrCGAN can generate compatible and realistic images under
compatibility learning setting compared to baselines.
While the difference against random images is small in the
Polyvore survey, MrCGAN is significantly preferred in the Amazon co-purchase survey. This
is also consistent with the discriminative performance.

\subsection{Discussion}
As shown in Table~\ref{tbl.fashion},
a larger $ \K $ gives better performance
when the total embedding dimension
$ (\K + 1) \times \N $ is kept equal,
but the differences are small if the size is increased continuously.
The regularizer controlled by $ \lambda_m $ forces the distances between prototypes and a compatible item to be small. We found that this constraint hurts recommendation performance, but it could be helpful for generation. We recommend choosing a smaller $ \lambda_m $ when the quality difference between images generated from $ \Ek{k}{x} $ and $ \Ek{0}{y} $ is not noticeable.

From the four datasets, the recommendation performance of our model is gradually getting closer to the others, which might be due to the disappearance of the asymmetric relationship. In Fashion-MNIST+1+2, this relationship is injected by force. Then Amazon also-bought/viewed dataset preserves buying and viewing orders so some asymmetric relationship exists. However, only the symmetric relationship remains for Amazon co-purchase and Polyvore datasets. This suggests our model is suitable for asymmetric relationship but it still works well under symmetric settings.

We found that a larger $ m_{prj} $ reduced the diversity but
a smaller $ m_{prj} $ made incompatible items more likely to appear.
In practice, it works well to set $ m_{prj} $ to be slightly
larger than the average distances of positive pairs from the training set.
Removing margin $ m_{enc} $ seems to decrease the diversity in simple datasets such as MNIST+1+2,
but we did not tune it on other complex datasets.

\section{Conclusion}\label{sec.con}
We propose modeling the asymmetric and many-to-many relationship of compatibility by learning
a Compatibility Family of representation and prototypes with an end-to-end system
and the novel Projected Compatible Distance function.
The learned Compatibility Family achieves more accurate recommendation results
when compared with the state-of-the-art Monomer method~\cite{HePacMcA16} on real-world datasets.
Furthermore, the learned Compatibility Family resides in a meaningful Compatibility Space
and can be seamlessly coupled with our proposed MrCGAN model to generate images of compatible items.
The generated images validate the capability of the Compatibilty Family in modeling many-to-many relationships.
Furthermore, when compared with other approaches for generating compatible images,
the proposed MrCGAN model is significantly more preferred in our user surveys.
The recommendation and generation results justify the usefulness of the learned Compatibility Family.

\section*{Acknowledgement}
The authors thank Chih-Han Yu and other colleagues of Appier as well as the anonymous reviewers for their constructive comments. We thank Chia-Yu Joey Tung for organizing the user study. The work was mostly completed during Min Sun's visit to Appier for summer research, and was part of the industrial collaboration project between National Tsing Hua University and Appier. We also thank the supports from MOST project 106-2221-E-007-107.

\fontsize{9.0pt}{10.0pt}
\selectfont
\bibliography{reference}
\bibliographystyle{aaai}
\end{document}